\documentclass[letterpaper, 10 pt, conference]{ieeeconf} 

\IEEEoverridecommandlockouts 

\usepackage{times} 
\usepackage{amsmath} 
\usepackage{amssymb} 
\usepackage{graphicx}
\usepackage{cases}
\usepackage{algorithm}
\usepackage{algorithmic}
\usepackage{caption}
\usepackage{color}
\usepackage{subfigure}
\usepackage{cite}

\title{\LARGE \bf
Robust shape estimation with false-positive contact detection
}

\author{Kazuki Shibata$^{1}$, Tatsuya Miyano$^{1}$, Tomohiko Jimbo$^{1}$ and Takamitsu Matsubara$^{2}$
\thanks{$^{1}$The authors are with the Autonomous Distributed Cooperative Control Program, Data Analytics Research-Domain, Toyota Central R$\&$D Labs., Inc., 41-1, Yokomichi, Nagakute, Aichi, Japan.
{\tt\small kshibata@mosk.tytlabs.co.jp}}%
\thanks{$^{2}$The authors are with Graduate School of Science and Technology, Division of Information Science, Nara Institute of Science and Technology, Nara, Japan
{}}%
}

\begin{document}

\maketitle
\thispagestyle{empty}
\pagestyle{empty}

\begin{abstract}
We propose a means of omni-directional contact detection using accelerometers instead of tactile sensors for object shape estimation using touch.
Unlike tactile sensors, our contact-based detection method tends to induce a degree of uncertainty with false-positive contact data because the sensors may react not only to actual contact but also to the unstable behavior of the robot.
Therefore, it is crucial to consider a robust shape estimation method capable of handling such false-positive contact data.
To realize this, we introduce the concept of heteroscedasticity into the contact data and propose a robust shape estimation algorithm based on Gaussian process implicit surfaces (GPIS).
We confirmed that our algorithm not only reduces shape estimation errors caused by false-positive contact data but also distinguishes false-positive contact data more clearly than the GPIS through simulations and actual experiments using a quadcopter.
\end{abstract}
\section{Introduction}

Capturing the shapes of objects is an essential issue in many areas of autonomous control such as infrastructure inspection and environmental monitoring.
In particular, to inspect a 3D construction whose shape is not available a priori, shape estimation is required for inspection path planning, using the data acquired from the available sensors.

\begin{figure}
\centering
\subfigure[3D construction]{
\includegraphics[width=7cm]{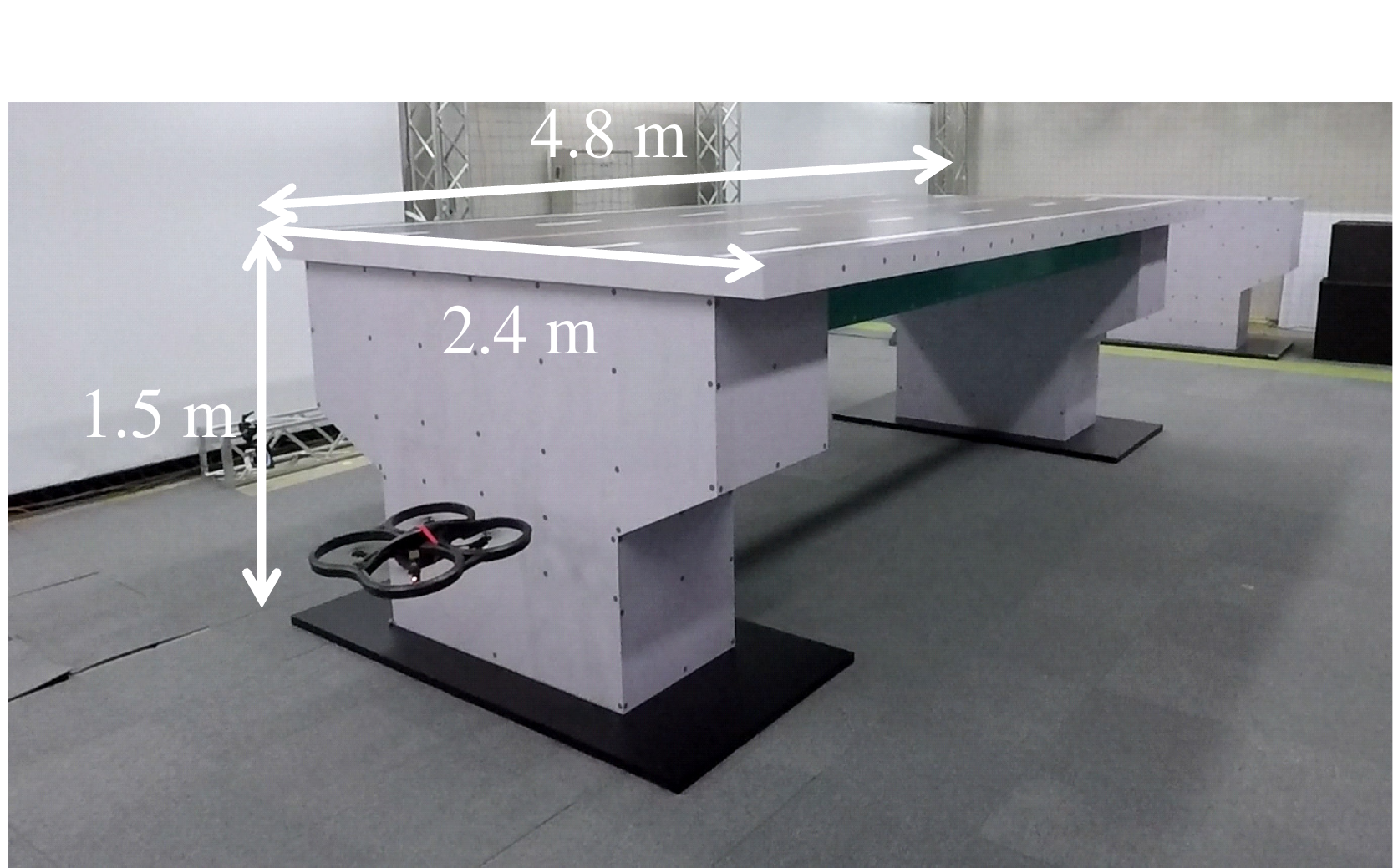}
\label{fig1a}}
\subfigure[Observed data and flight trajectory]{
\includegraphics[width=8cm]{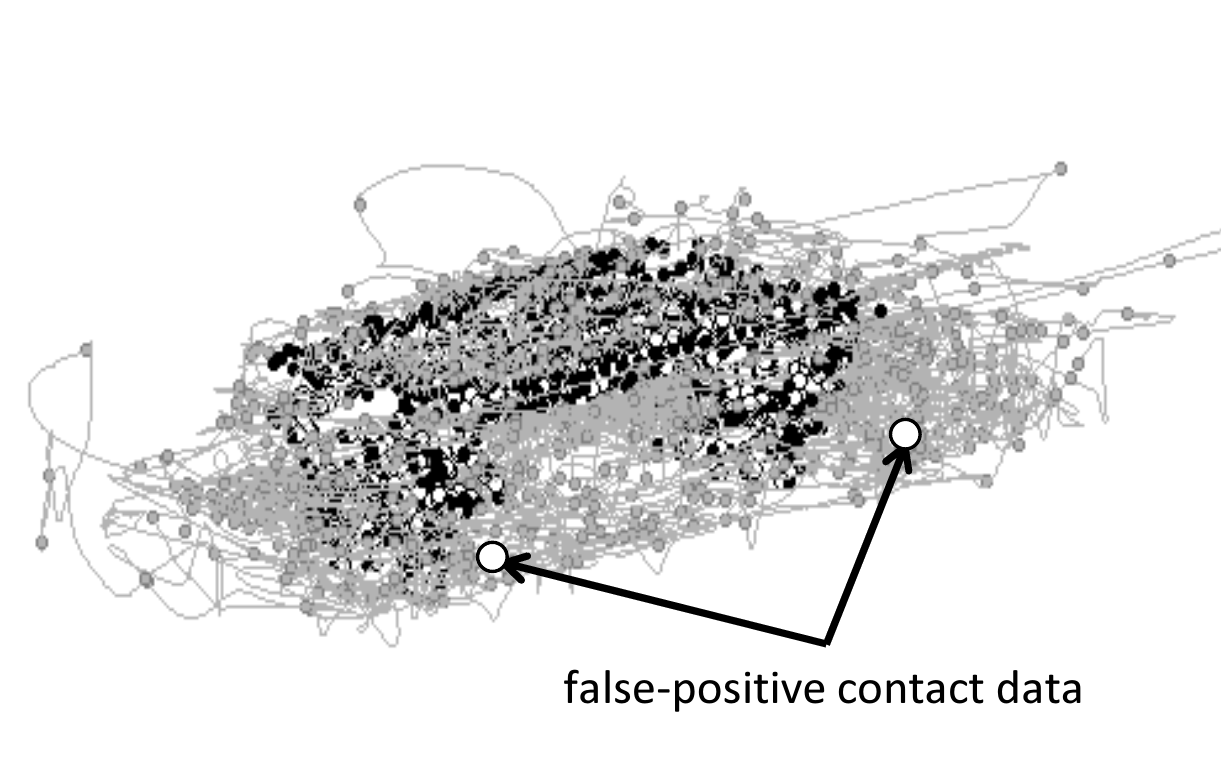}
\label{fig1b}}
\caption{(a) 3D construction. (b) Observed contact data and flight trajectory of quadcopter. The circles, black dots, gray dots, and gray lines represent internal points, contact points, external points, and the flight trajectory of the quadcopter, respectively.} 
\label{intro_fig}
\end{figure}

In this paper, we considered a large 3D construction which was subjected to estimation using a quadcopter, as shown in Fig. \ref{intro_fig}(a).
While most 3D reconstruction efforts using quadcopters have relied on vision sensors, they are noisy and suffer from occlusion.
In previous studies, tactile sensing was used to supplement the vision-based sensing and thus improve the shape estimation \cite{Ilonen, Luo, Bjorkman}.

Most touch-to-sense approaches, including grasping \cite{Saal, Li2014, Li2016} and manipulation \cite{Sanchez, Hoof, Chitta}, have explicitly assumed that contact occurrence is correctly detected by a tactile sensor. 
However, such an assumption may not always be realistic.
Given that it is difficult to determine, in advance, which parts of the robot will touch the construction, a number of sensors should be mounted on the quadcopter in an omnidirectional configuration. 
However, this may be unfeasible in terms of the robot's payload, cost, or power consumption. 

We explored an alternative approach to the estimation of the shape of an object by touch \it{without}\rm{} the need for tactile sensors. 
That is, we considered the use of an accelerometer that is commonly incorporated into a quadcopter to detect the occurrence of physical contact in any direction. 
This approach would avoid the issues associated with payload, cost, and power consumption. However, contact detection tends to be uncertain relative to when tactile sensors are being used. 
Unfortunately, the accelerometer may react not only to actual contact with the construction but also to unstable behavior induced by gusts of wind, collisions with other agents, etc.
As a result, false-positive contact data could be combined with normal contact data, as shown in Fig. \ref{intro_fig}(b).
Therefore, when using this approach, it is crucial to incorporate a robust shape estimation method capable of handling such false-positive contact data.
Note that, in the present study, we focused on false-positive contact data because false-negative contacts have little influence on the accuracy of the shape estimation.

Our approach to the estimation of the object shapes by means of touch without the use of tactile sensors involves the use of a robust shape estimation algorithm based on GPIS \cite{Williams, Caccamo, Mahler, Dragiev2011, Gerardo}, that is, robust GPIS.
While the GPIS assumes that the observations of contact data follow a normal distribution with a constant variance, our GPIS introduces heteroscedacity into each item of contact data.
Thus, we can prevent incorrect shape estimates produced by false-positive contact data.

To demonstrate the effectiveness of our approach, we constructed an experimental system using a quadcopter, using its built-in accelerometer for contact detection. We confirmed that our approach is better than GPIS for a 3D construction when using a quadcopter.

The contributions of the present study can be summarized as follows:

\begin{itemize}
\item the proposition of an approach for object shape estimation by touch with an omni-directional contact detection using accelerometers (without tactile sensors). 
\item the reduction in shape estimation errors caused by false-positive contact data.
\item the clear detection of false-positive contact data when applying our algorithm.
\end{itemize}

The remainder of this paper is organized as follows: 
Section II introduces related works, then Section III describes the shape estimation problem caused by false-positive contact data. 
Section IV introduces the mathematical formula for the GPIS and robust GPIS method, then Section V shows the effectiveness of our algorithm as demonstrated by numerous simulations. 
Section VI introduces the constructed experimental system and shows a few of the results of the experiments performed with the quadcopter. Finally, Section VII concludes the paper.

\section{Related work}
Previous studies addressing 3D reconstruction using quadcopters have usually used vision sensors \cite{Karrer, Teixeira, Vempati}.
Other studies have used lidar sensors \cite{Holz, Kumar} or a combination of vision and lidar sensors \cite{Michael, Shi}. 
However, these sensors sometimes cannot be used in harsh environments, in which diffused reflection occurs. 
To overcome this shortfall, several works \cite{Ilonen, Luo, Bjorkman} have employed tactle sensing to supplement the vision-based sensing.
Therefore, we considered a problem whereby we estimated a 3D construction by using a touch-to-sense approach.

Several recent studies have addressed shape estimation using deep learning.
Wu et al. \cite{Wu} and Dai et al. \cite{Dai} proposed the use of 3D-CNN for shape estimation, and
Varley et al. \cite{Varley} proposed a novel grasping framework based on the architecture in \cite{Dai}.
Watkins-Valls et al. \cite{Watkins} proposed an architecture that incorporates depth images and tactile information to improve shape estimation in occluded regions.
Through experiments using different types of 3D object shapes, they confirmed improvements in shape estimation errors and in the success rate of grasping compared with the GPIS.
While these studies obtained impressive performance in shape estimation, all except for Lundell et al. \cite{Lundell} are based on the deterministic model.
Therefore, it is not straightforward to apply this architecture to a model where there is estimation uncertainty and different kinds of noise.
In contrast, the GPIS handles this better by representing object shapes in a GP model.
Therefore, we focus on the use of GPIS to deal with false-positive contact data.

Other studies have focused on the use of GPIS for shape estimation using uncertain measurements obtained by multiple sensor modalities.
Caccamo et al. \cite{Caccamo} and Mahler et al. \cite{Mahler} combined visual data with haptic measurements using GPIS.
In addition, using the basic framework of GPIS, Dragiev et al. \cite{Dragiev2011} employed a combination of laser and tactile sensing while Gerardo-Castro \cite{Gerardo} used a combination of lasers and radar.

Basically, most of the GPIS methods are based on a homoscedastic GP, in which the distribution of observations is assumed to follow a normal distribution with a constant variance. However, this assumption may be unsuitable for application to cases involving some outliers because both normal and outlier data are handled equally.

To overcome this issue, several authors proposed heteroscedastic GPs (HGPs), for which the variation in the input or output noise is considered to be non-uniform.
Kuss et al. \cite{Kuss} and Jylanki et al. \cite{Jylanki} proposed a robust GP, which is a HGP that assumes that the noise variance is data-dependent.
McHutchon et al. \cite{McHutchon} proposed a GP with input noise (NIGP). This is a HGP that assumes that the noise variance differs for each input dimension.
Lazaro-Gredilla et al. \cite{Lazaro} proposed a variational heteroscedastic GP (VHGP), which is a HGP that assumes that the noise variance is input-dependent.
Among the available HGPs, we found that the robust GP is particularly suitable for our purpose given that it provides us with the uncertainty of data.

Among practical research using a robust GP as a mathematical model, Martinez-Cantin et al. \cite{MartinezC} proposed robust Bayesian optimization for active exploration of optimal policy parameters in a humanoid robot walking under the condition that there exist outliers of rewards caused by perturbations.
While our methodology used a common robust GP as a mathematical model, it differs from the methodology in \cite{MartinezC} in terms of shape reconstruction and false-positive contact detection when considering the uncertainty for each item of contact data.

\section{Touch-based object shape estimation using accelerometers}

This section introduces the shape estimation problem considered in the present study.
The goal of the present study was to estimate the shape of an unknown object placed in a given exploration region $Q\in \mathcal{R}^3$.

In \cite{Williams}, an implicit surface is used to describe the shape of an object by means of a real-value shape potential function. Based on this shape potential function, we can determine whether each point is located on the surface of, inside, or some distance from the object. 
As described in \cite{Dragiev2011}, we defined the observation of a shape potential value $y_i$ ($i=1,\cdots,n$) at the observed point $\textit{\textbf{x}}_i \in R^3$ as

$$
y_i=
  \begin{cases}
   0, \textit{\textbf{x}}_i \ \rm{on \ the \ surface}\\
   1, \textit{\textbf{x}}_i \ \rm{outside \ the \ surface}\\
   -1, \textit{\textbf{x}}_i \ \rm{inside \ the \ surface}.\\
  \end{cases}
\eqno{}
$$

In this study, the shape of the object is computed offline. Moreover, we assumed that the position of the quadcopter is accurate.
This assumption holds true when using high-precision positioning systems such as motion capture systems or RTK-GPS.

\begin{figure}
\begin{center}
\vspace{1mm}
\hspace{2mm}
\includegraphics[width=7.5cm]{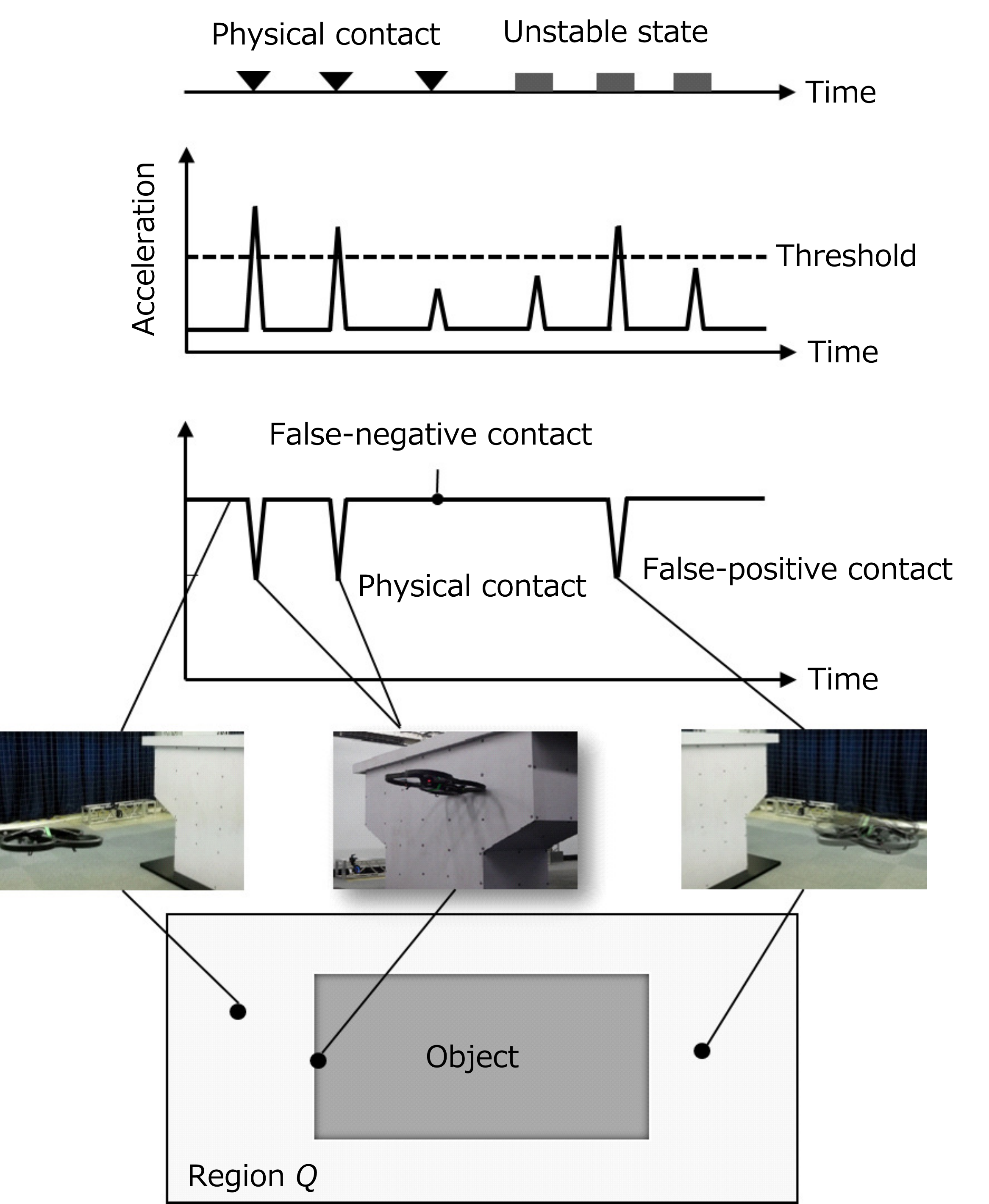}
\vspace{1.5mm}
\caption{Event detection and contact detection based on acceleration. When physical contact occurs, the acceleration becomes larger than a threshold (dashed line), such that the measurement $y$ will be zero. In this case, false-negative contact data can be detected when the quadcopter collides with an object at low speed.
Moreover, false-positive contact data are detected when the quadcopter becomes unstable and the acceleration exceeds the threshold.}\label{event_detection}
\end{center}
\end{figure}

\subsection{Observation model}

Observations were obtained by detecting events.
An illustration of this event detection when using an acceleration is shown in Fig. \ref{event_detection}.
In this example, an event takes the form of the quadcopter colliding with the object or becoming unstable.
If an event occurs and the acceleration becomes larger than the threshold, $y_i$ is labeled as being zero.
When physical contact occurs, false-negative contact data can occur when the quadcopter collides with the object at low speed.
Moreover, false-positive contact data are detected when the quadcopter becomes unstable.

As shown in Fig. \ref{event_detection}, it is difficult to determine the threshold so that the number of false-negative and false-positive contact data will be zero.
Although the number of items of false-positive contact data could be reduced by constructing more sophisticated classifiers, it is unrealistic to classify false-positive contact data with a 100 $\%$ accuracy rate for objects made of any material.
Therefore, in the present study, we explored another approach by assuming that the generation of false-positive contact data is unavoidable.

\subsection{Distribution of observations}

With GPs, we are required to assume the distribution of observations.
With homoscedastic GPs, the distribution is assumed to follow a normal distribution, as defined by
$$
\mathcal{N}(y_i\mid f_i,\sigma^2)=\frac{1}{\sqrt{2\pi \sigma^2}}\exp \left[ -\frac{(y_i-f_i)^2}{2\sigma^2} \right],
\eqno{}
$$
where $f_i$ and $\sigma^2$ represent the shape potential function and the noise variance with a constant value, respectively.
This distribution may be unsuitable for application to our problem; contact data with false-positive samples cannot be explained by such a light-tailed distribution, owing to the nature of the false-positive contact data and its representation with three natural numbers (-1/0/1). 
Thus, it may be unsuitable for use with false-positive contact data.

\begin{figure}
\begin{center}
\vspace{1mm}
\hspace{2mm}
\includegraphics[width=7.5cm]{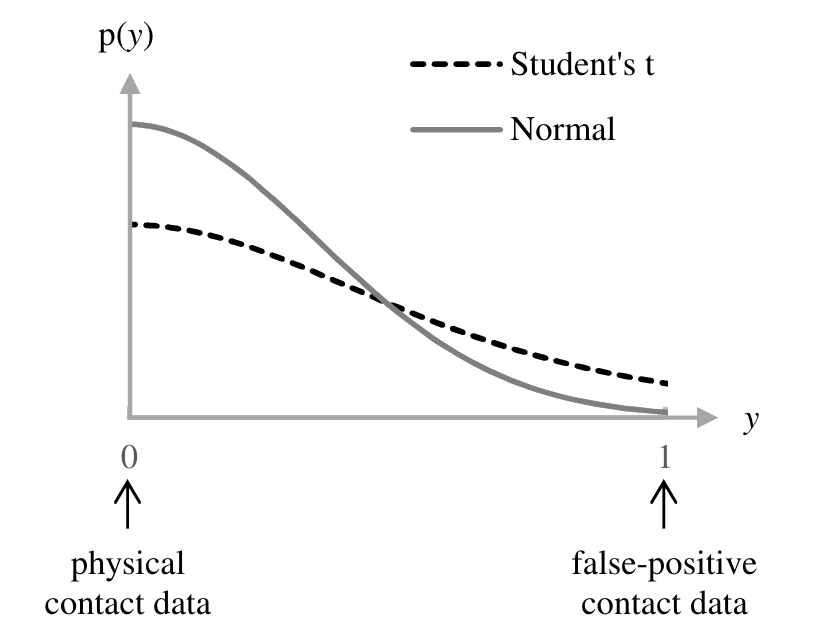}
\vspace{1.5mm}
\caption{Comparison of Student's t- and normal distribution.}\label{comparison_of_distribution}
\end{center}
\end{figure}

Therefore, we introduce the Student's t-distribution for the observations, given by
$$
\mathcal{T}(y_i\mid f_i,\lambda,\nu)=\frac{\Gamma(\frac{\nu+1}{2})}{\Gamma(\frac{\nu}{2})}\left( \frac{\lambda}{\pi \nu}\right)^{\frac{1}{2}}\left[1+\frac{\lambda(y_i-f_i)^2}{\nu} \right]^{-\frac{\nu+1}{2}},
\eqno{}
$$
where $\nu$, $\lambda$ and $\Gamma$ represent the degree of freedom, a scale parameter and gamma function, respectively.
As shown in Fig. \ref{comparison_of_distribution}, false-positive contact data can be explained with a greater degree of probability, owing to its heavy-tailed distribution.

\section{Methodology}

In this section, we describe our robust shape estimation with false-positive contact data.
We start with a brief introduction of GPIS, and then discuss the issues associated with the method.
Moreover, we propose a robust GPIS method by modifying the distributions of the observations made with the GPIS method.

\subsection{GPIS}

In a GPIS framework, the shape potential function $f$ is modeled by GP regression for a given data set $\mathcal{D}=\{\textit{\textbf{x}}_i, y_i \}$ ($i=1,\cdots,n$).
In the GPIS, a prior of the shape potential functions $\textit{\textbf{f}}:=[f(\textit{\textbf{x}}_1),\cdots,f(\textit{\textbf{x}}_n)]^{\rm{T}}{}$ is assumed to follow a normal distribution: $p(\textit{\textbf{f}}\mid \textit{\textbf{x}})=\mathcal{N}(\textit{\textbf{f}}\mid \textbf{0}, \textit{\textbf{K}})$, where $\textit{\textbf{x}}:=[\textit{\textbf{x}}_1,\cdots,\textit{\textbf{x}}_n]^{\rm{T}}{}$,
and $\textit{\textbf{K}} \in R^{n\times n}$ is the training data covariance matrix. 
Moreover, the probability of observations $\textit{\textbf{y}}:=[y_1,\cdots,y_n]^{\rm{T}}{}$ is given by the normal distribution with constant variance: that is, $p(\textit{\textbf{y}}\mid \textit{\textbf{f}})=\mathcal{N}(\textit{\textbf{y}}\mid \textit{\textbf{f}}, \sigma^2\textit{\textbf{I}})$.
Upon combining these probabilities, the posterior of the shape potential function is given by
\begin{eqnarray}
p(\textit{\textbf{f}}\mid \mathcal{D})=\frac{\mathcal{N}(\textit{\textbf{y}}\mid \textit{\textbf{f}},\sigma^2\textit{\textbf{I}})\mathcal{N}(\textit{\textbf{f}}\mid \textbf{0},\textit{\textbf{K}})}{p(\textit{\textbf{y}}\mid \textit{\textbf{x}})},
\label{shape_potential}
\end{eqnarray}
for which the marginal likelihood $\int \mathcal{N}(\textit{\textbf{y}}\mid \textit{\textbf{f}},\sigma^2\textit{\textbf{I}})\mathcal{N}(\textit{\textbf{f}}\mid \textbf{0},\textit{\textbf{K}})d\textit{\textbf{f}}$ in Eq. (\ref{shape_potential}) can be computed analytically as described in \cite{Rasmussen, Bishop}.
Moreover, the predictive mean and variance can be computed as
\begin{eqnarray}
\mu(\textit{\textbf{x}}^*)&=&\textit{\textbf{k}}^{\rm{T}}{}(\textit{\textbf{x}}^*)(\textit{\textbf{K}}+\sigma^2\textit{\textbf{I}})^{-1}\textit{\textbf{y}}, \label{GPmean} \\
\sigma^2(\textit{\textbf{x}}^*)&=&k(\textit{\textbf{x}}^*)-\textit{\textbf{k}}^{\rm{T}}{}(\textit{\textbf{x}}^*)(\textit{\textbf{K}}+\sigma^2\textit{\textbf{I}})^{-1}\textit{\textbf{k}}(\textit{\textbf{x}}^*), \label{GPvar}
\end{eqnarray}
where $k(\textit{\textbf{x}}_i, \textit{\textbf{x}}_j)$ is the covariance between the observed points $\textit{\textbf{x}}_i$ and $\textit{\textbf{x}}_j$, $i, j \in \{1, \cdots, n\}$,
$\textit{\textbf{k}}(\textit{\textbf{x}}^*)$ is the covariance vector between all the observed points $\textit{\textbf{x}}_i$ $(i=1,\cdots,n)$ and the test point $\textit{\textbf{x}}^*$, $\textit{\textbf{K}}$
has entries $K_{ij}=k(\textit{\textbf{x}}_i, \textit{\textbf{x}}_j)$, $i, j \in \{1, \cdots, n\}$.
An estimate of the object's shape is obtained by finding those points $\textit{\textbf{x}}^*$ where $\mu(\textit{\textbf{x}}^*)\approx 0$ in Eq. (\ref{GPmean}).

Because all the diagonal terms of $\sigma^2 \textit{\textbf{I}}$ in Eqs. (\ref{GPmean}) and (\ref{GPvar}) have the same value,
false-positive contact data have the same influence on the predictive distribution as normal data.
An issue, however, is whether the distribution of the observations is assumed to conform to a normal distribution with constant variance.

\subsection{Proposed method}

In this subsection, we propose a robust GPIS with false-positive contact data based on the GPIS method.

To realize this, we used the Student's t-distribution as the distribution of the observations, given by 
\begin{eqnarray}
p(\textit{\textbf{y}}\mid \textit{\textbf{f}})=\mathcal{T}(\textit{\textbf{y}}\mid \textit{\textbf{f}}, \lambda, \nu).
\label{distribution_of_observation}
\end{eqnarray}

Given the prior of the shape potential functions with a normal distribution, $p(\textit{\textbf{f}}\mid \textit{\textbf{x}})=\mathcal{N}(\textit{\textbf{f}}\mid \textbf{0}, \textit{\textbf{K}})$, the posterior of the shape potential function is given by
\begin{eqnarray}
p(\textit{\textbf{f}}\mid\mathcal{D},\boldsymbol{\psi},\lambda,\nu)=\frac{\mathcal{T}({\textit{\textbf{y}}\mid \textit{\textbf{f}}},\lambda,\nu)\mathcal{N}(\textit{\textbf{f}}\mid \textbf{0}, \textit{\textbf{K}})}{p(\textit{\textbf{y}}\mid \textit{\textbf{x}},\boldsymbol{\psi},\lambda,\nu)},
\label{posterior}
\end{eqnarray}
where $\boldsymbol{\psi}$ represents the hyperparameters in the covariance matrix $\textit{\textbf{K}}$.

In the same way as in the GPIS method, an estimate of the object's shape can be obtained from the posterior mean.
However, the marginal likelihood $\int \mathcal{T}({\textit{\textbf{y}}\mid \textit{\textbf{f}}},\lambda,\nu)\mathcal{N}(\textit{\textbf{f}}\mid \textbf{0}, \textit{\textbf{K}})d\textit{\textbf{f}}$ in Eq. (\ref{posterior}) cannot be computed analytically.
Thus, we conform to the variational approximation for GP regression with the Student's t-distribution, as described in \cite{Kuss}.

The Student's t-distribution in Eq. (\ref{distribution_of_observation}) is a scale-mixture of an infinite number of normal distributions, given by
\begin{eqnarray}
\mathcal{T}(y\mid f,\lambda,\nu)=\int^{\infty}_0\mathcal{N}(y\mid f,\sigma^2)\rm{Inv\Gamma}(\sigma^2\mid \alpha,\beta)d\sigma^2,
\label{Studentt}
\end{eqnarray}
where $\alpha=\nu/2$ and $\beta=\nu/{2\lambda}$ represent the shape and inverse scale parameter, respectively.
The Student's t-distribution becomes heavy-tailed when $\alpha$ becomes small or $\beta$ becomes large, as described in Appendix.
Since outliers are more likely to occur from a heavy-tailed distribution, it is suitable to explain false-positive contact data.
As described later, these parameters are optimized such that the estimated shape best fits the observed data.
Moreover, as $\alpha$ becomes small or $\beta$ becomes large, the noise variance, $\sigma^2$, becomes large with high probability, as described in Appendix.

Using Eq. (\ref{Studentt}), we approximate the posterior in the factorized form given by $p(\textit{\textbf{f}},\boldsymbol{\sigma}^2\mid \mathcal{D},\boldsymbol{\psi},\boldsymbol{\theta})\approx q(\textit{\textbf{f}})q(\boldsymbol{\sigma}^2)$,
where $\boldsymbol{\theta}:=[\alpha,\beta]^{\rm{T}}$ and $\boldsymbol{\sigma}^2=[\sigma_1^2,\cdots,\sigma_n^2]^{\rm{T}}$.
The approximate posterior of $\textit{\textbf{f}}$ and $\boldsymbol{\sigma}^2$ is given by
\begin{eqnarray}
q(\textit{\textbf{f}})&=&\mathcal{N}(\textit{\textbf{f}}\mid\textit{\textbf{m}},\textit{\textbf{A}}),\label{approximate_posterior_mean} \\
q(\boldsymbol{\sigma}^2)&=&\prod ^n_{i=1}\rm{Inv} \Gamma (\sigma^2_{\it{i}}{} \mid \tilde{\alpha}_\textit{i}{}, \tilde{\beta}_\textit{i}),\label{approximate_posterior_var}
\end{eqnarray}
where $\textit{\textbf{m}}$ and $\textit{\textbf{A}}$ represent the mean and covariance of $q(\textit{\textbf{f}})$, and 
$\tilde{\alpha}_i$ and $\tilde{\beta}_i$ represent the shape and inverse scale for each item of data $i\ (i=1,\cdots,n)$, respectively.
If we could approximate the posterior in an analytically tractable form, we could estimate an object's shape from its posterior mean.

To realize this, we introduce the KL divergence between the approximation and the posterior distribution, given by
\begin{eqnarray*}
\rm{KL}\it{}(q\| p)=\int q(\textit{\textbf{f}})q(\boldsymbol{\sigma}^2)\rm{ln}\it{}\frac{p(\textit{\textbf{f}},\boldsymbol{\sigma}^{\rm{2}\it{}}\mid \mathcal{D},\boldsymbol{\psi},\boldsymbol{\theta})}{q(\textit{\textbf{f}})q(\boldsymbol{\sigma}^2)}d\textit{\textbf{f}}d\boldsymbol{\sigma}^2.
\end{eqnarray*}
The KL divergence is minimized using an expectation maximization (EM) algorithm \cite{Dempster}.
In the EM algorithm, the approximate posterior and parameter values are updated by repeating the two steps.
In the expectation step (E-step), $\textit{\textbf{m}}$, $\textit{\textbf{A}}$ in Eq. (\ref{approximate_posterior_mean}), $\tilde{\alpha}_i$ and $\tilde{\beta}_i$ in Eq. (\ref{approximate_posterior_var}) are iteratively updated to minimize the KL divergence for given parameters $\boldsymbol{\theta}$ and $\boldsymbol{\psi}$.
In the maximization step (M-step), performed after each E-step, $\boldsymbol{\theta}$ and $\boldsymbol{\psi}$ are iteratively updated to minimize the KL divergence for fixed $q(\textit{\textbf{f}})$ and $q(\boldsymbol{\sigma}^2)$.
The EM algorithm iterates the E- and M-steps until the KL divergence converges.
Details of the method used to derive the update laws can be found in \cite{Kuss}.

After the variational approximation, the mean and variance of the approximate posterior are computed as
\begin{eqnarray}
\mu(\textit{\textbf{x}}^*)&=&\textit{\textbf{k}}^{\rm{T}}{}(\textit{\textbf{x}}^*)(\textit{\textbf{K}}+\boldsymbol{\Sigma})^{-1}\textit{\textbf{y}}, \label{robustGPmean}\\
\sigma^2(\textit{\textbf{x}}^*)&=&k(\textit{\textbf{x}}^*)-\textit{\textbf{k}}^{\rm{T}}{}(\textit{\textbf{x}}^*)(\textit{\textbf{K}}+\boldsymbol{\Sigma})^{-1}\textit{\textbf{k}}(\textit{\textbf{x}}^*), \label{robustGPvar}
\end{eqnarray}
where $\boldsymbol{\Sigma}$ represents a diagonal matrix with entries $\Sigma_{ii}=\tilde{\beta}_i/\tilde{\alpha}_i$ $(i=1,\cdots,n)$.
Since the diagonal terms of $\boldsymbol{\Sigma}$ in Eqs. (\ref{robustGPmean}) and (\ref{robustGPvar}) differ from each other, we can introduce heteroscedacity into each item of contact data.

The object shape is reconstructed by finding those points $\textit{\textbf{x}}^*$ that satisfy $\mu(\textit{\textbf{x}}^*)\approx 0$ in Eq. (\ref{robustGPmean}).
As $\tilde{\beta}_i/\tilde{\alpha}_i$ becomes large, the contact data $i$ has little influence on $\mu(\textit{\textbf{x}}^*)$.
Therefore, such contact data has little influence on the shape estimates.

Finally, we present the false-positive contact detection. After the variational approximation, we can obtain the uncertainty of data $i$, given by
\begin{eqnarray}
u_i=\tilde{\beta}_i/\tilde{\alpha}_i.
\label{uncertainty_of_data}
\end{eqnarray}
The noise variance, $\sigma^2$, in Eq. (\ref{Studentt}) becomes large with high probability as $u_i$ becomes large.
Therefore, we can judge the uncertainty for each item of contact data using Eq. (\ref{uncertainty_of_data}).

The main steps used in the shape estimation and false-positive contact detection are shown in Algorithm 1.

\begin{algorithm}[!tp]
\caption{: Robust GPIS}
\begin{algorithmic}[1]
\REQUIRE $x_i$, $y_i$ ($i=1,\cdots,n$),  $\boldsymbol{\theta}$, $\boldsymbol{\psi}$, $\textit{\textbf{x}}^*$
\ENSURE  $\textit{\textbf{x}}^*$ s.t. $\mu(\textit{\textbf{x}}^*)\approx0$, $u_i$  ($i=1,\cdots,n$)
\\ \textbf{Initialisation} : set $\tilde{\alpha}_i\leftarrow 1$, $\tilde{\beta}_i\leftarrow 1$, $\boldsymbol{\theta}$ and $\boldsymbol{\psi}$ to random positive values.
\STATE /* procedure for robust GP
\STATE Compute  $\textit{\textbf{K}}$ from  $\textit{\textbf{x}}$ and $\boldsymbol{\psi}$
\STATE E-step:
\STATE repeat
\STATE \ \ \ \ $\textit{\textbf{m}}$, $\textit{\textbf{A}}$, $\tilde{\alpha}_i$ and $\tilde{\beta}_i$ are iteratively updated for fixed
\STATE \ \ \ \ parameters $\boldsymbol{\theta}$ and $\boldsymbol{\psi}$ using Eqs. (5.34)-(5.36) in \cite{Kuss}.
\STATE until $\rm{KL}\it{}(q\| p)$ converges
\STATE M-step:
\STATE repeat
\STATE \ \ \ \ $\boldsymbol{\theta}$ and $\boldsymbol{\psi}$ are iteratively updated for fixed parameters
\STATE \ \ \ \ $\textit{\textbf{m}}$, $\textit{\textbf{A}}$, $\tilde{\alpha}_i$ and $\tilde{\beta}_i$ using Eqs. (5.39)-(5.43) in \cite{Kuss}.
\STATE until $\rm{KL}\it{}(q\| p)$ converges
\STATE Compute $\mu(\textit{\textbf{x}}^*)$ and $\sigma^2(\textit{\textbf{x}}^*)$ using Eqs. (\ref{robustGPmean}) and (\ref{robustGPvar}).
\STATE /* procedure for 3D shape reconstuction
\STATE Estimate object shape by finding those points $\textit{\textbf{x}}^*$ that satisfy $\mu(\textit{\textbf{x}}^*)\approx0$.
\STATE /* procedure for false-positive contact detection
\STATE Estimate the uncertainty of data $i$ from Eq. (\ref{uncertainty_of_data})
\end{algorithmic}
\end{algorithm}

\section{Simulation}

We conducted numerous simulations with 2D objects to confirm the reduction in the shape estimation errors and the false-positive contact detection when applying our algorithm.

\subsection{Experimental setup}
\begin{figure}
\centering
\subfigure[Square]{
\includegraphics[width=2.5cm]{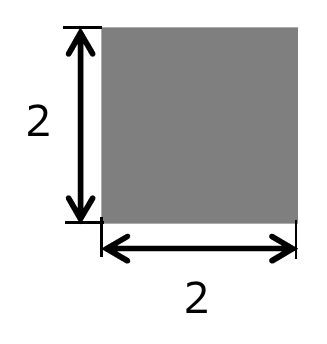}
\label{fig4a}}
\subfigure[Circle]{
\includegraphics[width=2.5cm]{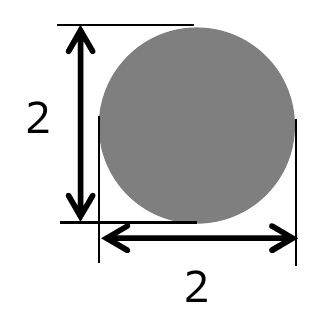}
\label{fig4b}}
\subfigure[Cross]{
\includegraphics[width=2.5cm]{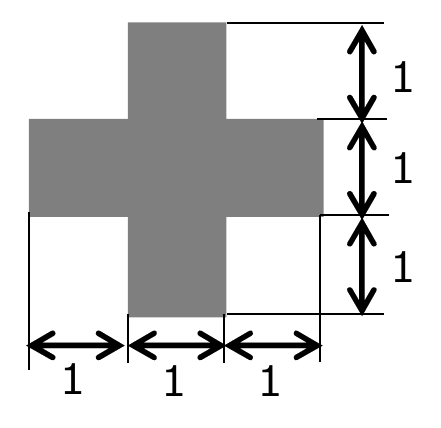}
\label{fig4c}}
\caption{2D object models used in simulation experiments. All fifures are in units of m.}
\label{shape2d}
\end{figure}

\begin{figure}
\centering
\subfigure[Square]{
\includegraphics[width=2.7cm]{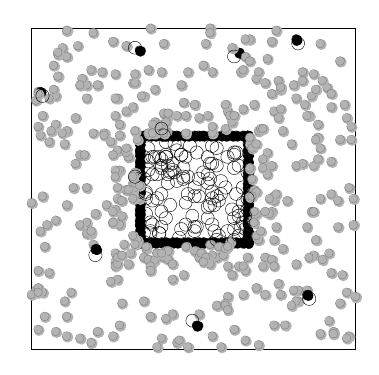}
\label{fig5a}}
\subfigure[Circle]{
\includegraphics[width=2.7cm]{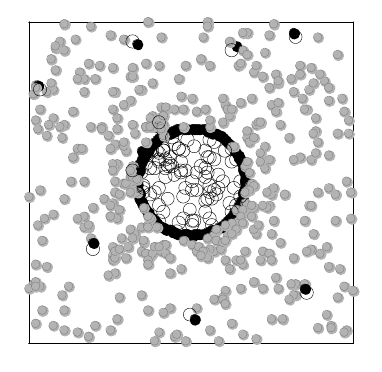}
\label{fig5b}}
\subfigure[Cross]{
\includegraphics[width=6.3cm]{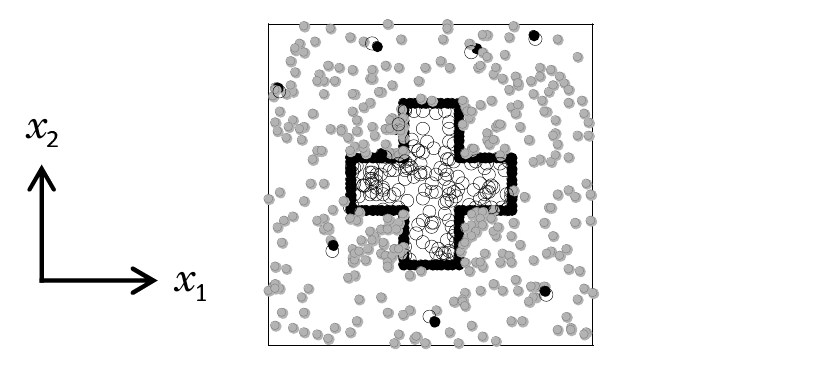}
\label{fig5c}}
\caption{Data acquired in one of the simulations. The circles, black dots, and gray dots represent internal points, contact points, and external points, respectively.}
\label{observed_data2d}
\end{figure}

We prepared three different object shapes, as shown in Fig. \ref{shape2d}.
The contact points were set on the edges at intervals of 0.01 [m], as shown in Fig. \ref{observed_data2d}(a) and (c), while they were set on the circumference at an interval of 3[deg] in Fig. \ref{observed_data2d}(b).
The internal and external points were randomly set in the given region $Q:=\{(x_1,x_2)\mid (-3\le x_1 \le 3, -3\le x_2 \le 3)\}$.
To consider the false-positive contact data, we replaced the external points with a pair of outliers of contact and internal points with a 2$\%$ chance.
The outlier of the internal point was randomly set within a certain distance 0.1 [m] from the outlier of the contact point.

We performed 50 simulations for each object shape.
The parameters used in the simulations are listed in Table I.
The parameters used in the GPIS methods were set by trial and error.
The test data were set on a grid with an interval of 0.02 [m].



\begin{table}
\caption{Simulation conditions}
\renewcommand{\arraystretch}{1.5}
  \centering
  \begin{tabular}{ccc}
    \hline  
  Variable  & Symbol  &  Value  \\
    \hline
    Number of observed points & $n$ & 5.9$\times10^2$ \\    
    Number of test points & $n_t$ & 9.1$\times10^4$ \\ 
    Initial value of variance of Gaussian kernel  & - & $0.25^2$ \\
    Initial value of shape parameter  & $\alpha$ & 2.0  \\
    Initial value of inverse scale  & $\beta$ & 4.0 \\  
    \hline    
  \end{tabular}
\end{table}

\subsection{Shape estimation error}

\begin{figure*}
\centering
\subfigure[Square]{
\includegraphics[width=14cm]{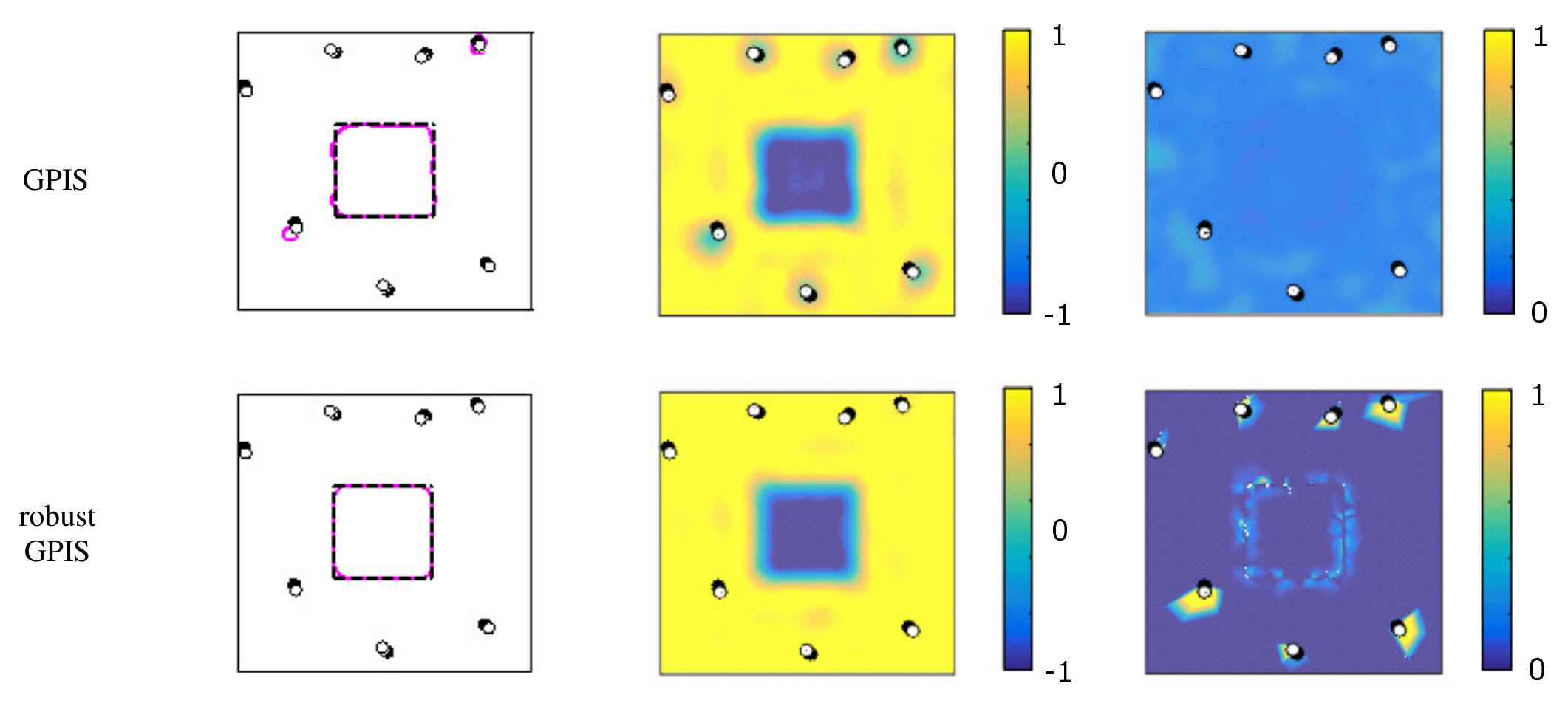}
\label{fig6a}}
\subfigure[Circle]{
\includegraphics[width=14cm]{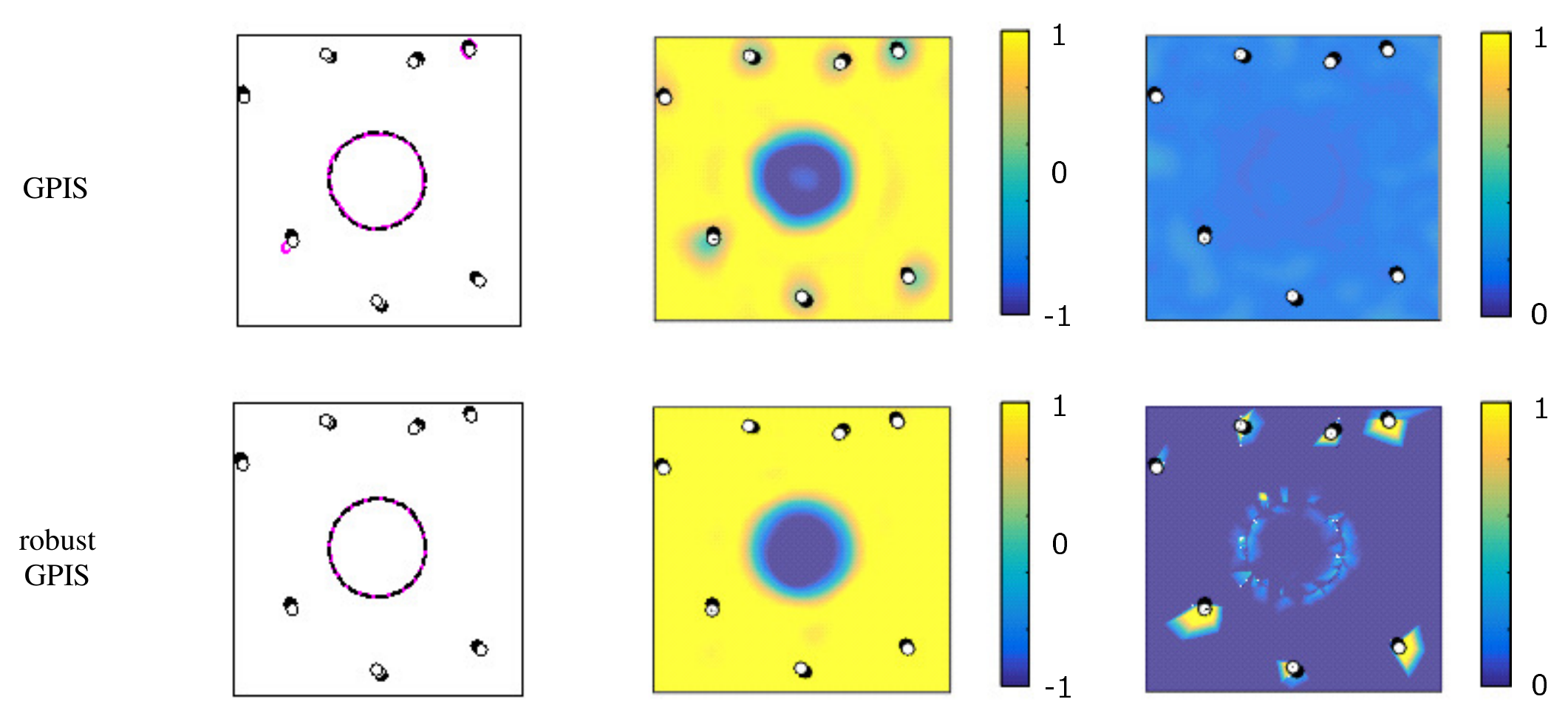}
\label{fig6b}}
\subfigure[Cross]{
\includegraphics[width=14cm]{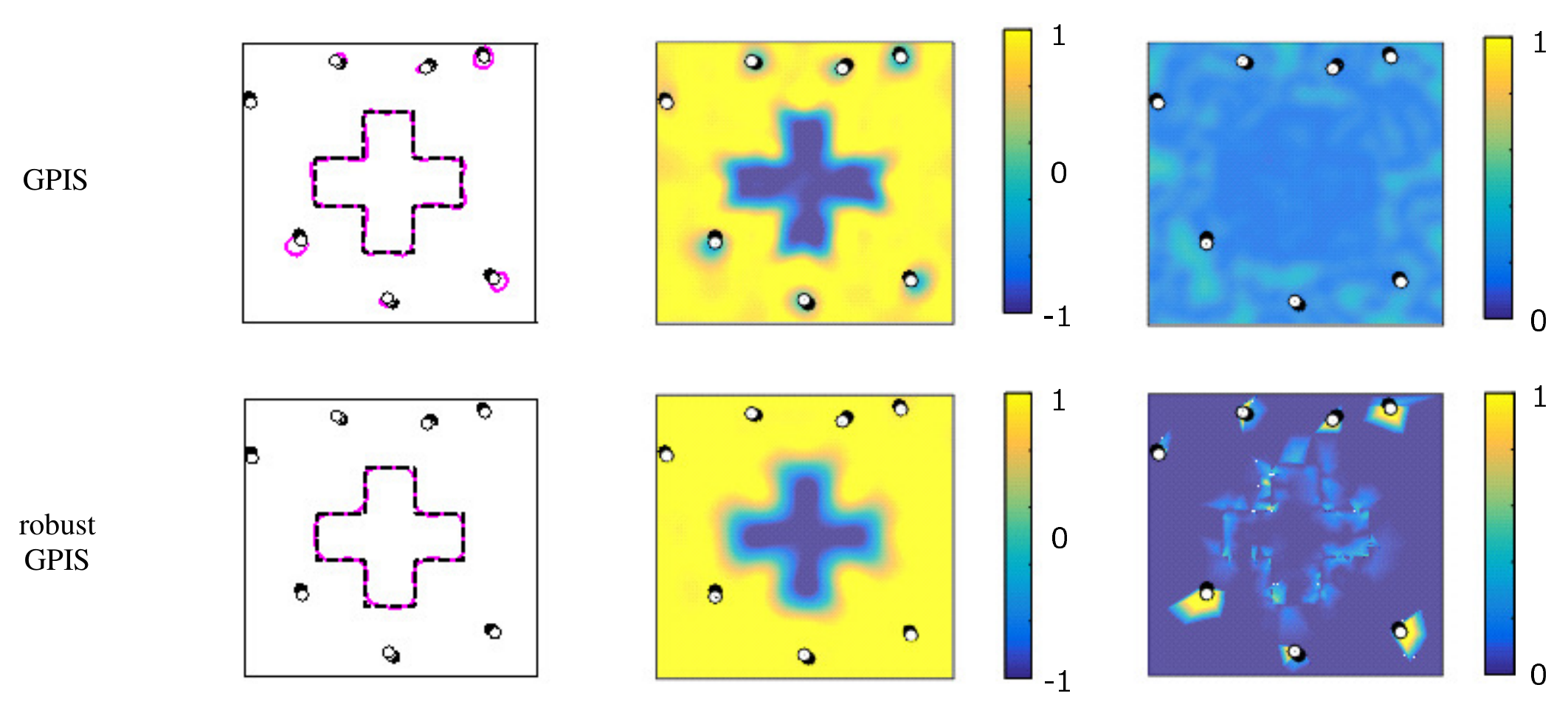}
\label{fig6c}}
\caption{Estimated shapes of 2D objects (left), predicted mean (center), and uncertainty of data (right) when applying the GPIS methods. 
The dashed lines, black dots, and circles represent the real object shapes, outliers of contact points, and outliers of internal points, respectively.}
\label{result2d}
\end{figure*}

One of the 50 simulations is shown in Fig. \ref{result2d}.
As shown in the figure, the object shapes could be roughly captured using the GPIS.
At the same time, however, surfaces were incorrectly estimated around a few of the outliers, located sufficiently far from the object.
The predicted shape potential values (colored contours) show that the mean values became negative around the outliers, which indicates that the predictive distributions were greatly influenced by the outliers.



In contrast, a surface was not incorrectly estimated around the outliers when using the robust GPIS, as shown in Fig. \ref{result2d}.
Considering the uncertainty values, those values around the outliers became remarkably large compared with those around the normal data.
As a result, the predicted shape potential values around the outliers were sufficiently greater than zero, indicating that the outliers had little influence on the predicted mean values.

To evaluate the accuracy of shape estimation using the GPIS methods, we introduce metrics for determining the shape estimation error (i.e., the difference between the estimated shape and the true shape).
Using Eqs. (2) and (9), we selected the test point $i=\{1,\cdots, n_s \}$ whose mean value $\mu(\textit{\textbf{x}}_i)$ is almost equal to zero ($-0.01<\mu(\textit{\textbf{x}}_i)<0.01$), where $\textit{\textbf{x}}_i$ is the position of the test point $i$. 
After computing the minimum distance $d_i$ between the test point $i$ and the surface of the true shape, we computed the mean error of the distance $d_i$, defined by
\begin{eqnarray}
e=\frac{1}{n_s}\sum^{n_s}_{i=1}d_i.
\label{shape_error}
\end{eqnarray}

The results obtained with the t-test for the 50 simulations used to compare the robust GPIS with the GPIS are listed in Table II.
These results reveal significant differences between the robust GPIS and the GPIS in terms of the shape estimation errors, since the p-values are all less than 0.01.
Therefore, for all the prepared objects, our algorithm produced smaller shape errors than the GPIS.

\subsection{False-positive contact detection}

This subsection shows the clear detection of false-positive contact data when applying our algorithm. With the robust GPIS, the uncertainty of the data around the outliers became remarkably large compared with those around the normal data, as shown in Fig. \ref{result2d}.

Here we introduce metrics for evaluating the false-positive contact detection.
While the robust GPIS can estimate the uncertainty of data, the GPIS cannot directly do so.
Therefore, we regard the predicted variance at each observed data as the uncertainty of the data.
To define the clarity of the false-positive contact detection for each outlier $o$, we define the metrics given by
\begin{eqnarray}
Q_o=\frac{1}{n_o}\sum_{j\in \mathcal{R}_o}u_j\bigg/u_o,
\label{false_positive_contact_detection}
\end{eqnarray}
where $u_o$, $\mathcal{R}_o$ and $n_o$ represent the uncertainty of the data $o$, the neighborhood region within certain distance $d_o$ from the outlier $o$, and the number of the normal data within the region $\mathcal{R}_o$, respectively.
In the simulations, we set $d_o=0.50$ [m], which is equal to the length and width of the quadcopter used in the experiment.

Similar to the evaluation of the shape estimation error, we performed a t-test for the 50 simulations by comparing the robust GPIS with the GPIS.
The results listed in Table III reveal significant differences between the robust GPIS and the GPIS in terms of false-positive contact detection, since the p-values are all less than 0.01.

Therefore, for all the prepared objects, our algorithm could detect false-positive contacts more clearly than the GPIS.

\begin{table}
\caption{Results of t-tests comparing the GPIS and the robust GPIS with respect to shape estimation error. The degree of freedom is 98 in all cases. ($\ast \ast$: $p<0.01$)}
\renewcommand{\arraystretch}{1.5}
  \centering
  \begin{tabular}{c|c}
    \hline
     & p-value \\
    \hline
    Square & 7.7E-11$^{\ast \ast}$  \\
    Circle  & 7.7E-9$^{\ast \ast}$ \\
    Cross  & 1.2E-9$^{\ast \ast}$ \\
    \hline  
  \end{tabular}
\end{table}

\begin{table}
\caption{Results of t-tests comparing the GPIS and the robust GPIS with respect to false-positive contact detection. ($\ast \ast$: $p<0.01$)}
\renewcommand{\arraystretch}{1.5}
  \centering
  \begin{tabular}{c|c}
    \hline
     & p-value  \\
    \hline
    Square & 1.4E-104$^{\ast \ast}$ \\
    Circle  & 2.5E-111$^{\ast \ast}$ \\
    Cross  & 8.3E-121$^{\ast \ast}$ \\
    \hline  
  \end{tabular}
\end{table}

\section{Experiment}

In this section, we will demonstrate the use of the algorithm through experiments using a quadcopter, specifically, a Parrot AR.Drone 2.0.
We introduce the experimental configuration and conditions, as well as the omni-directional contact detection using the accelerometers implemented in the experiment. We then present the experimental results.

\subsection{Experimental configuration and conditions}


The position of the quadcopter was observed using an OptiTrack Prime 17W motion-capture system (Natural Point, Inc., Corvallis, OR) operating at 200 Hz.
In the experiment, the ground-truth shape of the 3D object is obtained by placing markers at the vertexes of the polygons that compose the object surfaces. 
Moreover, the object shape is considered in the ground-fixed coordinate system.
Therefore, translation and rotation were not considered when comparing the estimated shape with the true object shape.

The acceleration and attitude were measured using mounted sensors operating at 200 Hz.
The quadcopter was controlled using a personal computer (PC) with an 8-core Intel Core i7 (2.80 GHz) processor, 32 GB of RAM, and a joystick (Extreme 3D Pro, Logitech, Inc., Lausanne, Switzerland).
The measurement data and control signals were exchanged between the PC and the quadcopter via WiFi communication.

\clearpage

The exploration region $Q$ is given by
$$
\textit{\textbf{x}}\in Q:= \{\textit{\textbf{x}} \mid -4\le x_1\le 4, -2\le x_2\le 2, 0\le x_3\le 2 \},
\eqno{}
$$
where $x_1$, $x_2$, and $x_3$ represent the longitudinal, transverse, and vertical axes, respectively.

We controlled the quadcopter using the joystick, which allowed us to collect as much contact data as possible in $Q$ at 25 Hz while collecting non-contact data as uniformly as possible at 0.5 Hz.
The limitations imposed by the battery capacity forced us, for each experiment, to perform five flights of 8 min each.

To validate the effectiveness of our algorithm, we generated some ground truth outliers by applying a gust of wind in some regions.
We conducted two flights with artificial pertubations and prepared two observed data sets combined with the other four flights without artificial pertubations.
The initial values for the hyperparameters used with the GPIS methods are listed in Table IV.

\begin{table}
\caption{Experimental conditions}
\renewcommand{\arraystretch}{1.5}
  \centering
  \begin{tabular}{ccc}
    \hline  
  Variable  & Symbol  &  Value  \\
    \hline
    Threshold of contact detection [G]  & - & 0.6 \\
    Thickness parameter [m]  & $d_{in}$ & 0.1 \\
    Number of observed points & $n$ & 2.7$\times10^3$ \\    
    Number of test points & $n_t$ & 5.4$\times10^5$ \\ 
    Initial value of variance of Gaussian kernel  & - & $0.25^2$ \\
    Initial value of shape parameter  & $\alpha$ & 2.0  \\
    Initial value of inverse scale  & $\beta$ & 4.0 \\  
    \hline    
  \end{tabular}
\end{table}

\subsection{Omni-directional contact detection using accelerometers}

This subsection introduces omni-directional contact detection based on the acceleration of the quadcopter.
Based on the measured acceleration, the shape potential value is obtained using

$$
y_i=
  \begin{cases}
   0, \rm{if \ } \| \it{}^B\hat{\textit{\textbf{a}}}_i \|> \it{a}_{\rm{0}}\\
   \\
   1, \rm{otherwise}\\
  \end{cases},
\eqno{}
$$
where $^B \hat{\textit{\textbf{a}}}_{\textit{i}}$ and $a_0$ represent the acceleration of the robot in the body-fixed coordinate system $B$ and the positive threshold to detect the transition state from non-contact to contact, respectively.

\begin{figure}
\begin{center}
\vspace{1mm}
\hspace{2mm}
\includegraphics[width=8cm]{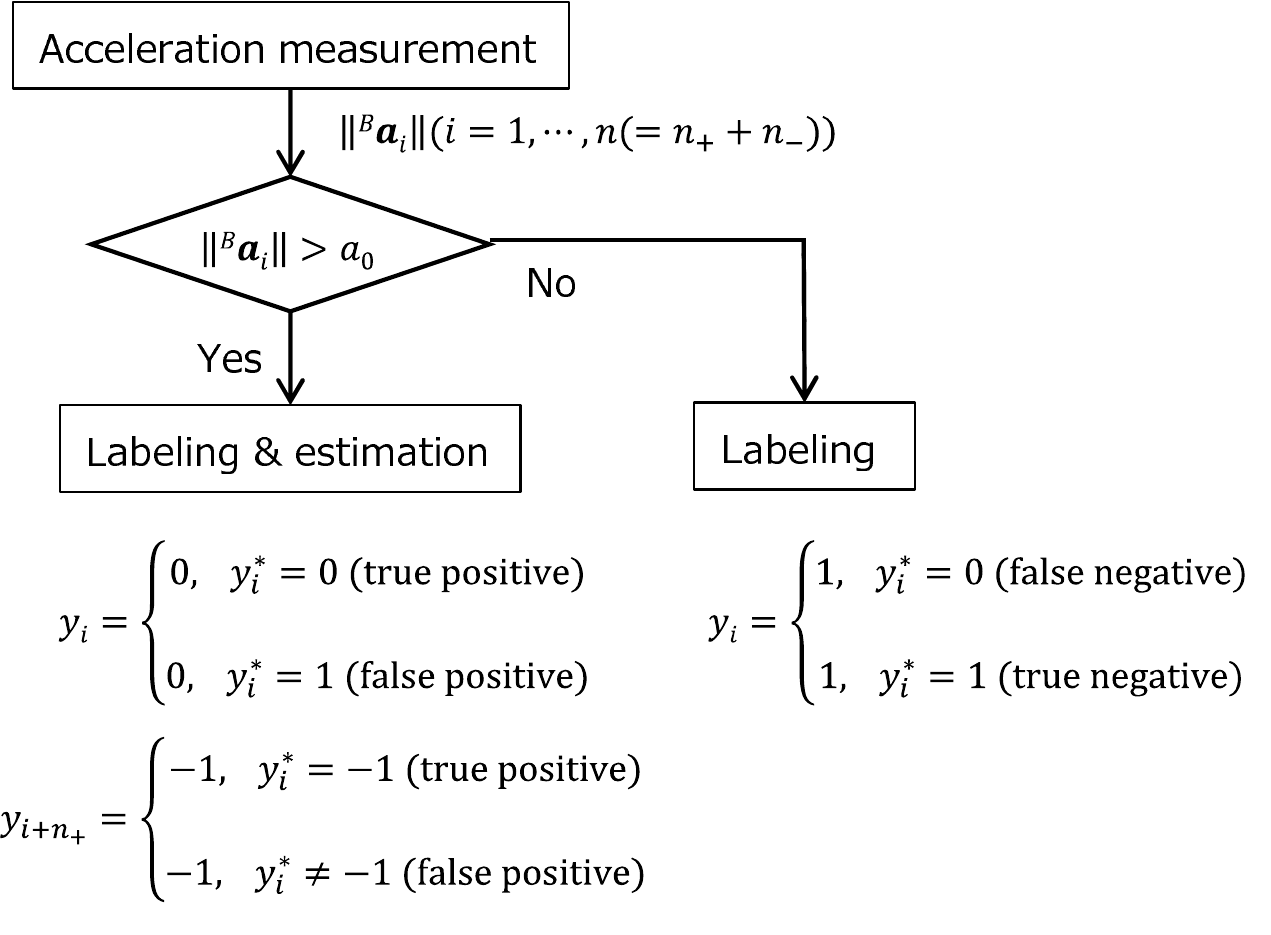}
\vspace{1.5mm}
\caption{Flowchart of omni-directional contact detection. $n_{+}$ and $n_{-}$ represent the number of contact and external points and the number of internal points, respectively.}\label{flowchart}
\end{center}
\end{figure}

The flowchart of the contact detection is shown in Fig. \ref{flowchart}. 
If $y_i$ is equal to zero and its true value $y_i^{\ast}$ is equal to one, the data $i$ is the false-positive contact data. 
If $y_i$ is equal to one and $y_i^{\ast}$ is equal to zero, the data $i$ is the false-negative contact data. 
The occurence of these data depend on the the threshold $a_0$.
If $a_0$ is set too low, false-positive contact detection can occur easily since sudden acceleration and deceleration can be detected as false-positive contacts.
In the opposite case, false-negative contact detection can occur.
In this paper, we focus on the false-positive contacts since the false-negative contacts have much less influence on the false-positive contacts. 
We searched for the minimum value of $a_0$ by trial and error, considering that the number of false-positive contacts is as small as possible. 

Furthermore, the outward normal vector on the surface is estimated by

\begin{eqnarray}
\hat{\textit{\textbf{n}}}_i=
  \begin{cases}
   \frac{^E\hat{\textit{\textbf{a}}}_i}{\| ^E\hat{\textit{\textbf{a}}}_i \|}, \rm{if \ } \it{}\it{y}_i=\rm{0} \\
   \\
   \textbf{0}, \rm{otherwise}\\
  \end{cases},
\label{normal_vector}
\end{eqnarray}
where $^E \hat{\textit{\textbf{a}}}_i$ represents the acceleration of the robot in the ground fixed coordinate system $E$.
Acceleration with mounted sensors is observed in the body-fixed coordinate system $B$, such that $^E \hat{\textit{\textbf{a}}}$ in Eq. (\ref{normal_vector}) is converted from $^E\hat{\textit{\textbf{a}}}_i=\textit{\textbf{R}}(\hat{\phi}) \textit{\textbf{R}}(\hat{\theta}) \textit{\textbf{R}}(\hat{\psi}) ^B\hat{\textit{\textbf{a}}}_i$,
where $\textit{\textbf{R}}$, $\hat{\phi}$, $\hat{\theta}$, and $\hat{\psi}$ represent the rotation matrix, and the measured values of the roll, pitch, and yaw angle, respectively.

\begin{figure}
\begin{center}
\vspace{1mm}
\hspace{2mm}
\includegraphics[width=7.5cm]{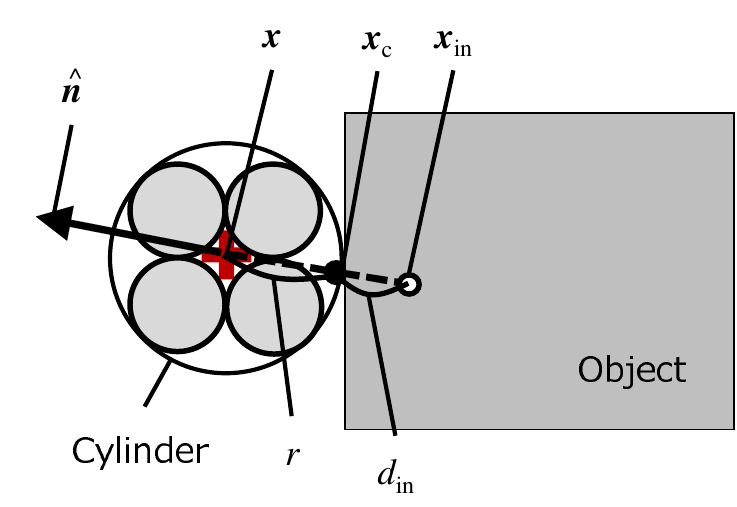}
\vspace{1.5mm}
\caption{Estimation of contact and internal points. When the robot collides with the object, a contact point (colored dot) is estimated using the center of gravity (cross), the estimated vector (arrow) and the circumscribed cylinder to the robot.
The internal point (circle) is estimated on the extension line through the center of gravity of the robot and the contact point.
}\label{contact_localization}
\end{center}
\end{figure}

Using the center of gravity of the robot and the estimated normal vector, we approximated a contact position $\textit{\textbf{x}}_c$, as shown in Fig. 8.
Consider a circumscribed cylinder to the quadcopter with the radius $r$ and the height $h$.
If one of the planar components of the estimated vector $\hat{\textit{\textbf{n}}}_i$ is largest, the contact position is calculated by $\textit{\textbf{x}}_{c_{i}}=\textit{\textbf{x}}_i-r\hat{\textit{\textbf{n}}}_i$.
Otherwise, the contact position is calculated by $\textit{\textbf{x}}_{c_{i}}=\textit{\textbf{x}}_i-h\hat{\textit{\textbf{n}}}_i$.

Because the internal points of the object cannot be obtained directly from the contact, we estimate the position of the internal point $\textit{\textbf{x}}_{in}$, as shown in Fig. 8.
Only if $y(\textit{\textbf{x}})=\rm{0}$, the position of the internal point and its potential value are calculated as $\textit{\textbf{x}}_{in}=\textit{\textbf{x}}_c-d_{in}  \hat{\textit{\textbf{n}}}(\textit{\textbf{x}})$ and $y(\textit{\textbf{x}}_{in})=-1$,
where $d_{in}$ represents the thickness parameter determined by the minimum thickness of the object.
The training data was produced by merging the contact, external, and internal points while the test data was set on a grid with an interval of 0.05 [m].

\subsection{Shape estimation error}

\begin{figure*}
\centering
\subfigure[GPIS]{
\includegraphics[width=7.2cm]{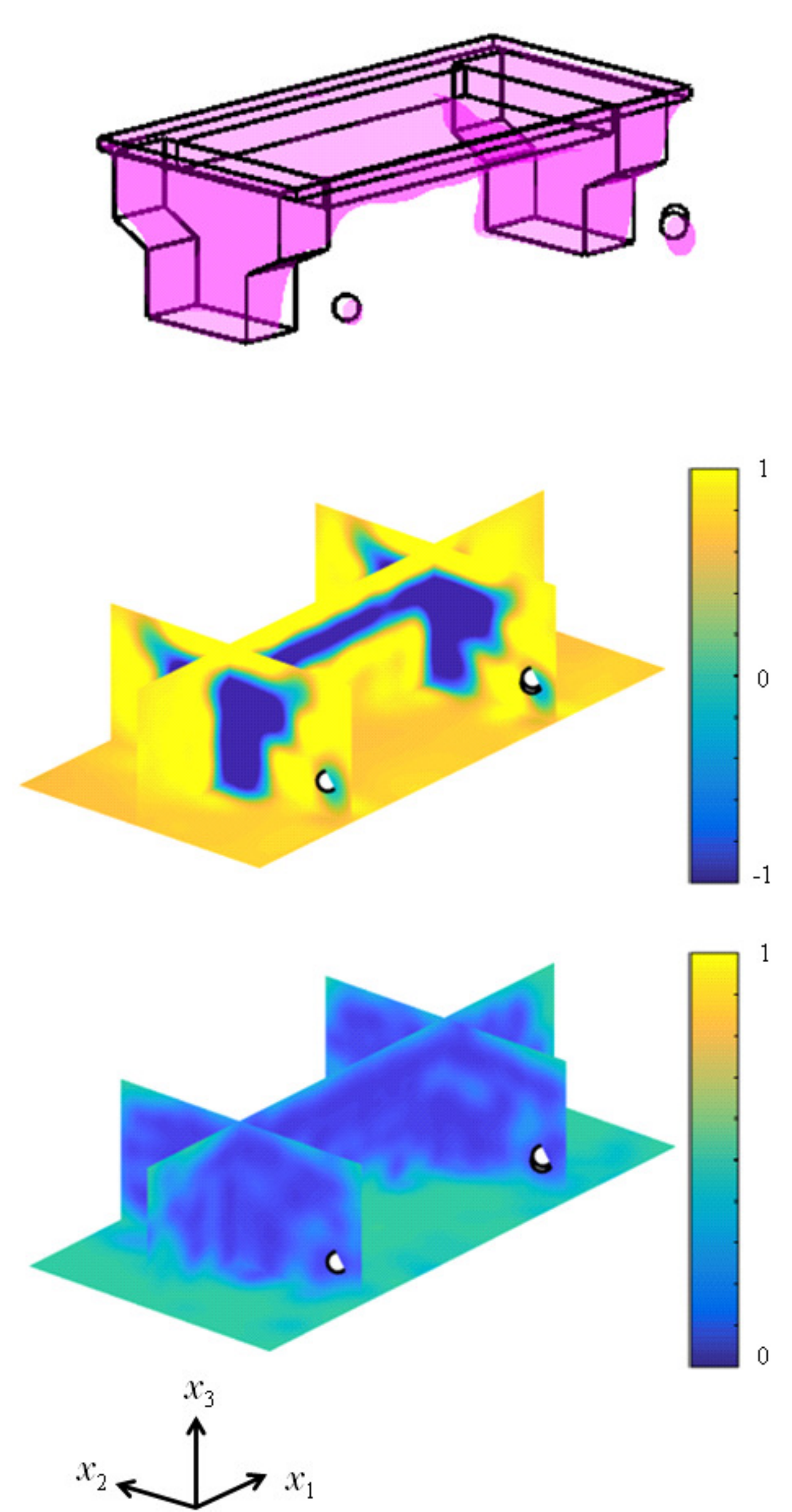}
\label{GPIS3d}}
\subfigure[robustGPIS]{
\includegraphics[width=7.2cm]{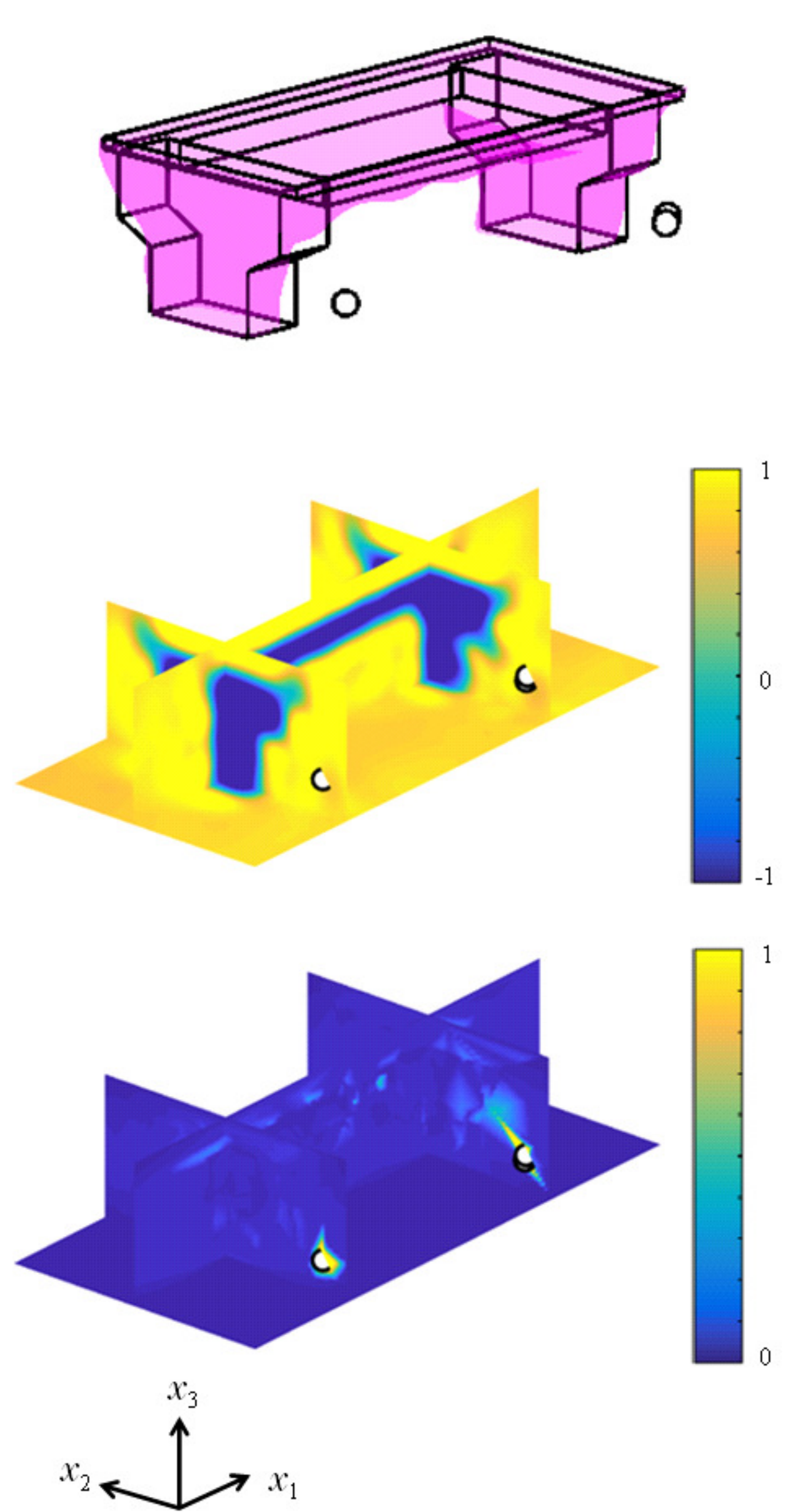}
\label{robustGPIS3d}}
\caption{Estimated shapes (top), predicted mean (middle), and uncertainty of data (bottom) when applying the GPIS methods to a 3D construction. The circles and bold lines represent the false-positive contact data and the true object shape, respectively. The cross-sections are placed in order to visualize the distribution of each value.}
\label{result3d}
\end{figure*}

The results of shape estimation when applying the GPIS methods are shown in Fig. \ref{result3d}.

As shown in Fig. \ref{result3d}(a), the shape of the 3D construction could be roughly captured when applying GPIS.
However, surfaces were generated at points sufficiently far from the construction.  
The predicted mean values of the shape potential function were all zero around the outliers.
In contrast, the robust GPIS was able to avoid the incorrect generation of surfaces around the outliers, as shown in Fig. \ref{result3d}(b).
This result indicates that the predicted mean values of the shape potential function were sufficiently greater than zero around the outliers.

Furthermore, the shape estimation errors computed by Eq. (\ref{shape_error}) when applying the GPIS methods are listed in Table V.
The results indicate that the use of the robust GPIS incurred smaller errors than the GPIS.

\subsection{False-positive contact detection}
This subsection focuses on the results of false-positive contact detection when applying the GPIS methods.

The distribution of the uncertainty of data, shown in Fig. \ref{result3d}(a), indicates that the uncertainty around the outliers was as small as that around the normal data with the application of GPIS.
In contrast, we can confirm that the uncertainty around the outliers in Fig. \ref{result3d}(b) could be distinguished very clearly when our algorithm was applied. 

To evaluate the false-positive contact detection of the GPIS methods, we evaluated false-positive contact detection for each outlier using Eq. (\ref{false_positive_contact_detection}).
In the experiment, we regarded any detected contact points which were 0.5 meters or more from the object as being outliers. This value was equivalent to the length and width of the quadcopter.

The mean values for eight outliers in the two experiments are listed in Table VI.
The results show that the mean value obtained when applying the GPIS were almost equal to one, which indicates that the GPIS could not distinguish between the outliers and the normal data.
In contrast, the mean values were much smaller than one, which indicates that our algorithm could very clearly distinguish the outliers from the normal data.

\begin{table}
\caption{Mean error of the estimated shapes}
\renewcommand{\arraystretch}{1.5}
  \centering
  \begin{tabular}{c|ccc}
    \hline
    Algorithm & GPIS  &  Robust GPIS  \\
    \hline
    Experiment\#1 & 0.10 m& \textbf{0.07} m\\
    Experiment\#2 & 0.12 m& \textbf{0.07} m\\
    \hline    
  \end{tabular}
\end{table}

\begin{table}
\caption{False-positive contact detection}
\renewcommand{\arraystretch}{1.5}
  \centering
  \begin{tabular}{c|ccc}
    \hline
    Algorithm & GPIS  &  Robust GPIS  \\
    \hline
    Mean values & 0.96 & \textbf{0.045} \\
    \hline   
  \end{tabular}
\end{table}

\subsection{Discussion}

In this subsection, we discuss the usefulness of the false-positive contact detection.
Consider active object exploration \cite{Yi, Dragiev2013, Martinez, Matsubara} with false-positive contact data.
In this problem, the outliers might degrade any prediction, resulting in inefficient exploration since robots might excessively explore some areas around the incorrect surfaces.
Moreover, false-negative contacts could degrade the efficiency of the object exploration although they were not considered in the present study.
If contacts were not detected at the surface, the GPIS methods could predict that no surface exists around the false-negative contacts.
As a result, those locations were difficult to sample when applying Bayesian optimization \cite{Mockus}.
To deal with this issue, we can remove the outliers based on our false-positive contact detection.
To apply our algorithm to active object exploration, we should reduce the amount of observed data using a sparse Gaussian process regression methods as described in \cite{Snelson, Titsias}.

Our future work will aim to improve the contact localization in our experimental system.
In a more complex scenario such as some concave formations, multiple contacts might occur at different locations on the quadcopter.
In such a situation, contacts are detected around the inside part of the corners because the current system estimates only one contact position at any one time using the resultant force of multiple contacts. 
As a result, the estimated shape around the corner tends to be rounded.
To overcome this issue, more precise contact localization should be implemented using the methods described in \cite{McMahan, Molchanov}.
In addition, we should extend our contact detection to handle multiple contacts, as proposed by Sommer et al. \cite{Sommer}.

In this study, we assumed that the position of the quadcopter is accurate.
However, in real environments, it is necessary to consider the uncertainty of localization in the shape estimation.
To address this issue, we aim to eliminate the assumption by combining the GP model with the input noise proposed by Girard et al. \cite{Girard}.


It would be interesting to combine our contact detection with sensor modalities other than an accelerometer.
Moreover, we could explore applications other than shape estimation when using vision sensors.

\section{Conclusion}

In this paper, we have proposed an approach for object shape estimation based on touch with omni-directional contact detection using accelerometers.
Because our contact detection method tends to induce a degree of uncertainty due to the presence of false-positive contact data, we proposed a robust shape estimation method capable of handling such false-positive contact data.
We confirmed that our algorithm could reduce shape estimation errors caused by false-positive contact data through simulations and actual experiments using a quadcopter.
Moreover, our algorithm could distinguish false-positive contact data more clearly than the GPIS, suggesting that false-positive contact detection which is possible with our proposed algorithm can be useful for other applications, such as active object exploration. 



\clearpage

\appendix

\section{Student's t- and inverse gamma distribution}
This appendix introduces how the parameters $\alpha$ and $\beta$ affect the Student's t- and inverse gamma distributions in Eq. (\ref{Studentt}).
The Student's t-distribution for different combinations of $\alpha$ and $\beta$ is shown in Fig. \ref{Studentt_distribution}.
Based on the results, the Student's t-distribution gets heavy-tailed when $\alpha$ becomes small or $\beta$ becomes large.

The inverse gamma distribution is given by
\begin{eqnarray*}
\rm{Inv}\Gamma\it{}(\sigma^2\mid \alpha,\beta)=\frac{\beta^{\alpha}}{\rm{\Gamma}\it{}(\alpha)}(\sigma^2)^{-(1+\alpha)}\exp \left(-\frac{\beta}{\sigma^2}\right).
\end{eqnarray*}
The inverse gamma distribution for different combinations of $\alpha$ and $\beta$ values is shown in Fig. \ref{inv_gamma_distribution}.
From the result, the noise variance, $\sigma^2$, becomes large with high probability as $\alpha$ becomes small or as $\beta$ becomes large.

\renewcommand{\thefigure}{A.1} 

\begin{figure}[!hp]
\begin{center}
\includegraphics[width=8cm]{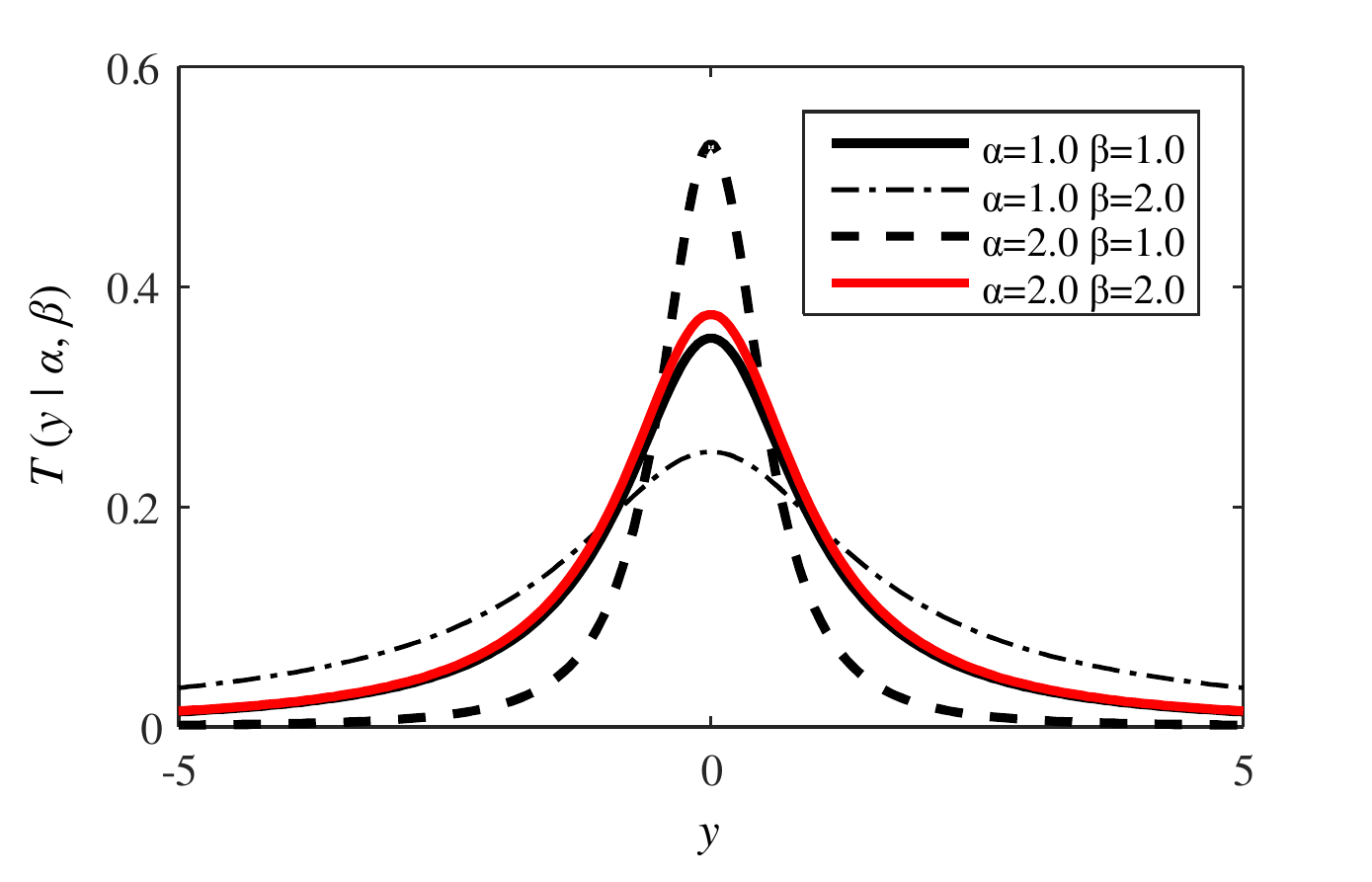}
\caption{Student's t-distribution}
\label{Studentt_distribution}
\end{center}
\end{figure}

\renewcommand{\thefigure}{A.2} 

\begin{figure}[!hp]
\begin{center}
\includegraphics[width=8cm]{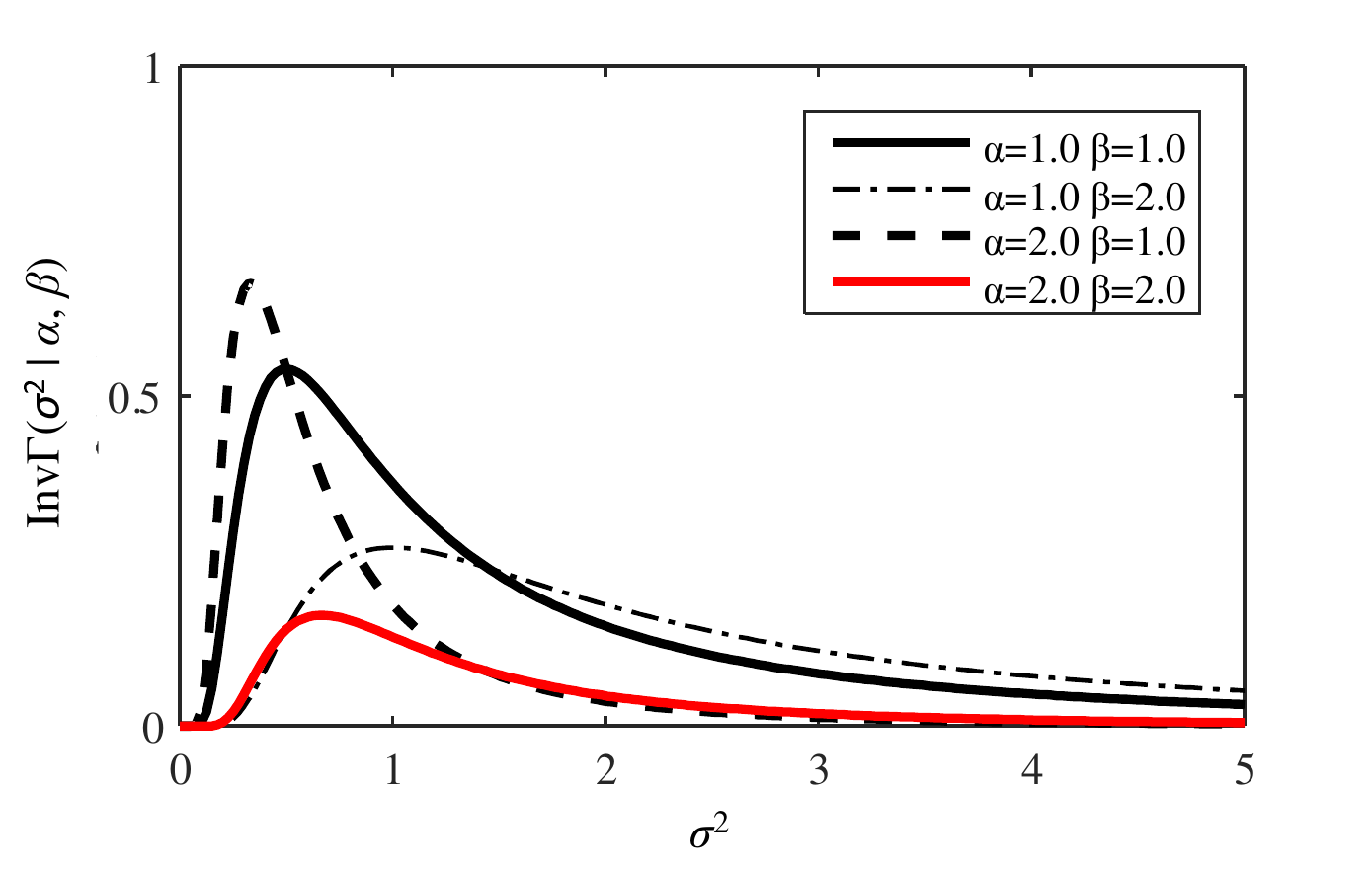}
\caption{Inverse gamma distribution}
\label{inv_gamma_distribution}
\end{center}
\end{figure}

\end{document}